\documentclass[runningheads]{llncs}
\usepackage[T1]{fontenc}
\usepackage{amsmath}
\usepackage{graphicx,verbatim}
\begin{document}
\title{VLM-KG: Multimodal Radiology Knowledge Graph Generation}
%
\author{Abdullah Abdullah\orcidID{0000-0003-0843-0112} \and
Seong Tae Kim \orcidID{0000-0002-2132-6021}}
\authorrunning{A. Abdullah and S. T. Kim}
%
\institute{Department of Computer Science and Engineering, Kyung Hee University, Republic of Korea \\
\email{\{abdullahijaz,st.kim\}@khu.ac.kr}}

\maketitle              
\begin{abstract}
Vision-Language Models (VLMs) have demonstrated remarkable success in natural language generation, excelling at instruction following and structured output generation. Knowledge graphs play a crucial role in radiology, serving as valuable sources of factual information and enhancing various downstream tasks. However, generating radiology-specific knowledge graphs presents significant challenges due to the specialized language of radiology reports and the limited availability of domain-specific data. Existing solutions are predominantly unimodal, meaning they generate knowledge graphs only from radiology reports while excluding radiographic images. Additionally, they struggle with long-form radiology data due to limited context length. To address these limitations, we propose a novel multimodal VLM-based framework for knowledge graph generation in radiology. Our approach outperforms previous methods and introduces the first multimodal solution for radiology knowledge graph generation. 


\end{abstract}
\section{Introduction}

Radiology data consists of free text reports containing critical information about a patient’s health based on an interpretation of radiology images and the patient's history. The unstructured nature of these radiology reports poses a challenge to downstream applications in the medical domain. Several strategies have been developed for automatic extraction of structured information. These methods range from using data labelers \cite{c3,r3,c2} and others that try to extract more fine-grained information in the radiology report \cite{r4}. Knowledge graphs in radiology provide a structured representation for these unstructured radiology reports and are a great source of factual information. \cite{r2} is the first work that proposed a schema-based extraction method for extracting information in the form of entities and relations. They proposed RadGraph, which is a knowledge graph of entities and relations from full-text radiology reports. RadGraph consists of 500 MIMIC-CXR radiology reports \cite{r6} annotated according to their schema by board-certified radiologists. Additionally, they developed a BERT \cite{c9} based model Dygiee++, which was trained on the RadGraph dataset to automatically generate entity and relation labels across 220,763 MIMIC-CXR  reports. In this schema, entities are identified as a continuous stretch of text containing one or more adjacent words that focus on anatomy and observation. A single relation connects two entities. RadGraph uses three relations, namely: suggestive\_of, located\_at, and modify. 
Several studies have utilized this dataset in tasks such as radiology report generation and pathology classification \cite{r13,r1,r28}.

However, RadGraph has major limitations. The annotations are limited to radiology reports, and the schema does not capture the clinical context of the radiology reports. Secondly, it fails to label the ambiguous cases in the medical reports due to the ambiguity of natural language. Moreover, the smaller context length of the model is a limitation in the generation of knowledge graphs for long-form radiology reports. 

Recent developments in large language models (LLMs) have shown impressive text-generation abilities. These text models are now extended to other modalities, such as visual inputs. Many solutions have been provided to bridge the gap between text and vision modality. \cite{r15} categorizes vision language models (VLMs) into four categories: contrastive, masked, pretrained, and generative VLMs. Our data which contains triplets consisting of entities and relations that are repeated across different reports in the dataset, makes it difficult to train a contrastive VLM. 
The masking-based VLMs require huge computation resources. Similarly, generative VLMs are also expensive to train. VLMs based on pretrained LLM backbones often leverage a projector to learn a mapping between the vision model and the LLM and are not resource-intensive to train. Therefore, in our work, we use a pretrained open-source LLM along with a contrastive vision model MedCLIP \cite{r11}, which is trained on radiology images. More details on the projector and our framework are provided in subsequent sections.

\subsubsection{Contribution}
1) To the best of our knowledge, it is the first study to introduce a multimodal knowledge graph generation framework for radiology. 2) Our VLM-based models, with their extended context length, achieve superior performance compared to previous radiology-specific solutions and demonstrate promising results for knowledge graph generation.

\begin{figure}
\includegraphics[width=\textwidth]{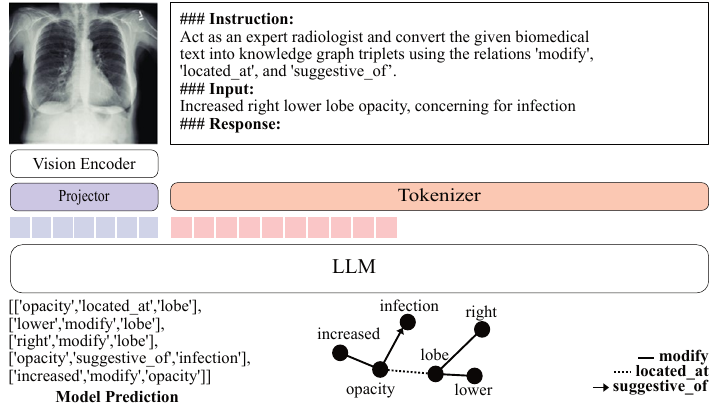}
\caption{Overview of visual instruction tuning based multimodal radiology knowledge graph generation framework. The instruction dataset format is shown(Top). The Projector and the LLM are trained on the visual input and the instruction data for multimodal knowledge graph generation.} \label{fig1}
\end{figure}

\section{Methodology}

\subsection{Instruction Tuning}

To leverage the capabilities of our pretrained LLM, we train our LLM backbone with an instruction tuning method. The format of the instruction dataset can be seen in Figure \ref{fig1} where the instruction is a prompt to the LLM, the input values are the radiology reports, and the output values are corresponding triplets. We refer to this Instruction tuned LLM as LLM-KG. It has learned to represent the radiology reports in the style of entities and relations. Instruction tuning enables our LLM to learn the information at the global level from the data and also the structure of the knowledge graph. 

\subsection{Visual Instruction Tuning}

To enable the LLM to work with medical images, we perform visual instruction tuning \cite{r17}. 
We design the framework with LLM, the vision model of MedCLIP~\cite{r11}, and the projector for learning to map the visual embeddings to the language embeddings. Both the projector and the language model are trained for downstream tasks. 
In the training, the output embeddings from the projector are concatenated with the instruction token embeddings from instruction data. The cross-entropy loss \cite{r16} is calculated, and backpropagation is performed to update the weights of the projector and the LLM-KG. We refer to this visually instruction-tuned LLM-KG as VLM-KG.

\subsubsection{Projector.}
Inspired by the works of \cite{r9}, we employ a compact 8-layer transformer-based architecture as a projector to learn to map visual information from MedCLIP's embedding space to the LLM-KG embedding space. This projector has a small number of parameters and is trained to effectively translate from MedCLIP's 512 dimensions to our LLM-KG's 1024 dimensions. A linear layer reshapes the image feature vector into a sequence format. The length of the sequence is based on the provided parameter named `clip\_length(k)'. A learned constant called prefix is added to the sequence, enriching it with additional, adjustable context. The length of this prefix is given by `prefix\_length(n)'. The prefix retrieves meaningful information from MedCLIP embedding through the multi-head attention, and it learns to adjust to the LLM-KG. The final output is tailored for compatibility with LLM-KG, focusing on the transformed sequence corresponding to the original image features. The output embeddings from the projector are then concatenated with the instruction token embeddings and these input embeds are passed to the LLM-KG which processes it and gives the output prediction. In the choice of projector, the MLP layer is widely used to bridge the pretrained LLMs with pretrained vision models. In these models, the MLP layer serves the purpose of translating the shape of embeddings from the vision model to the input embedding dimension of the LLM. But in our case, the projector not only requires to learn the translation of shape from the vision model to LLM but also the important embeddings for the radiology images. Therefore a more sophisticated projector is used as compared to an MLP layer.

\section{Experiments and Results}
\subsection{Datasets}

We train our framework on knowledge graph triplets from MIMIC-CXR \cite{r6} and IU-Xray datasets \cite{r29} generated from the RadGraph benchmark Dygiee++ model \cite{r2}.  
The training and validation set of the MIMIC-CXR knowledge graph triplets dataset consists of 230,558 and 1907 samples, respectively. The training and validation set of the IU-Xray knowledge graph triplets dataset contains 2,069 and 296 samples, respectively. 50 MIMIC-CXR radiology reports, which are annotated by board-certified radiologists, are used for evaluating models as a test set as in \cite{r2}.

\subsection{Training Details}

In our framework, we utilise the pretrianed Qwen1.5-0.5B LLM \cite{r19}. The Qwen1.5-0.5B model is used in this study because it is under 500 million parameters while it supports a context length of 32K tokens, which is much higher than the previous BERT-based models. This enables it to process long-form radiology reports, which was a limitation of previous models. Moreover, we wanted to demonstrate the instruction-following capabilities of the smaller LLMs and how they can be augmented in clinical settings where computational resources are low. We use MedCLIP, which is a contrastive model trained on the radiology images and reports from the MIMIC-CXR dataset. For our choice of projector, we utilize the transformer-based architecture. We trained the LLM on the instruction dataset by keeping a batch size of 2 for five epochs and with a gradient accumulation value of 2. The learning rate was set at 1e-4. We used the SFTTrainer from TRL library \cite{c6} for instruction tuning of our model. For visual instruction tuning, we trained the projector along with the LLM-KG with a batch size of 2 and a learning rate of 2e-5, along with a value of 5000 for warmup steps for five epochs. The AdamW \cite{r8} optimizer was incorporated. It took 40 hours to train for one epoch on 4 NVIDIA RTX A6000 GPUs. In general, our models output valid format. However, to make the responses consistent and for accurate evaluation, we perform post-processing. We convert the response string into a list, each representing a knowledge graph triplet using regular expressions.

\subsection{Evaluation Metrics}
We evaluate the effectiveness of our models on the task of knowledge graph triplets generation. The list of triplets is processed and converted to strings where punctuation, like commas and semicolons, are separated. The knowledge graph triplets generation is the task of natural language generation, and hence we employ Natural Language Generation (NLG) metrics such as BLEU \cite{c8}, ROUGE-L\cite{c7} scores to evaluate the performance of our models on this task.
The BLEU score compares n-grams in the generated text to those in a reference text. The following equation represents the BLEU score, where \(\text{BP}\) is the brevity penalty, \(N\) is the maximum \(n\)-gram order, and \(p_n\) is the precision for \(n\)-grams: 
\(\text{BLEU} = \text{BP} \cdot \exp \left( \frac{1}{N} \sum_{n=1}^{N} \log p_n \right)\). ROUGE-L measures the quality of text generation based on the longest common subsequence between the generated text and the reference text. The following equation represents the ROUGE-L score, where \(\text{RLCS}\) is the longest common subsequence score, \(\text{R}\) is recall, \(\text{P}\) is precision, and \(\beta\) balances the importance of precision and recall: 
\(\text{ROUGE-L} = \frac{(1 + \beta^2) \cdot \text{RLCS}}{\text{R} + \beta^2 \cdot \text{P}}\).

\subsection{Comparative Results}


\begin{table*}[t]
\caption{The table shows comparison of our models vs Radgraph Benchmark model Dygiee++ \cite{r2} on MIMIC-CXR reports annotated by expert radiologists for the task of knowledge graph triplets generation. LLM-KG refers to the instruction-tuned LLM. VLM-KG refers to the visual instruction-tuned LLM-KG. IU and MIMIC are the knowledge graph datasets used for training. ``B\_n" refers to n-gram BLEU scores,  and ``RL"  to ROUGE-L score.}
 \centering
 \begin{tabular}{c|cc|cc|ccccc}
 \hline
 &    \multicolumn{5}{c}{NLG metrics}\\
 \hline
 &      B1& B2& B3& B4&RL\\
 \hline
 Dygiee++&   24.57& 24.22& 23.81& 23.35&\underline{56.49}\\
 \hline
 LLM-KG(IU)&   25.32& 21.04& 18.02& 15.79&40.68\\
 VLM-KG(IU)& \underline{44.76}& \underline{38.84}& 34.84& 31.69&49.66\\
 LLM-KG(MIMIC)& 40.52& 37.91& \underline{35.96}& \underline{34.35}&\textbf{56.53}\\
 VLM-KG(MIMIC)&  \textbf{54.98}& \textbf{49.65}&\textbf{46.12}&\textbf{43.29}&54.69\\
 \hline
 \end{tabular}
  \label{tab1}
 \end{table*}

Table \ref{tab1} presents the performance of different models across standard NLG metrics, highlighting the impact of dataset size, multimodal learning, and domain adaptation. VLM-KG (MIMIC) achieves the best results across BLEU metrics, with BLEU-1 (54.98), BLEU-2 (49.65), BLEU-3 (46.12), and BLEU-4 (43.29), significantly outperforming other models due to its ability to leverage both visual and linguistic information, which enhances contextual understanding. Its ROUGE-L score (54.69), though slightly lower than LLM-KG (MIMIC) (56.53), suggests that while multimodal learning improves precision, it may introduce slight variations in generated text. The LLM-KG (MIMIC) model also shows strong performance, significantly exceeding LLM-KG (IU) across all metrics (e.g., BLEU-3: 35.96 vs. 18.02), demonstrating the benefits of training on a large-scale domain-specific dataset. 

\begin{figure}[t]
\includegraphics[width=\textwidth]{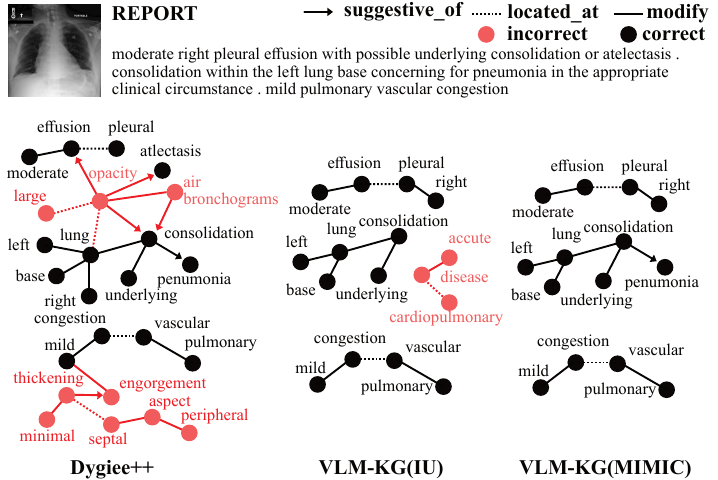}
\caption{Comparision of our multimodal VLM-KG models with unimodal Dygiee++ model for the task of knowledge graph generation. In the case of our models, both the radiographic image and the report in instruction format as shown in Figure \ref{fig1} are passed for the knowledge graph triplets generation. In the case of Dygiee++, which is an unimodal BERT-based model, only the radiology report is passed. The black color highlights the correct triplets corresponding to the radiology report. The red-colored triplets are incorrect and hallucinated predictions.} \label{fig2}
\end{figure}

The MIMIC dataset (about 230,000 samples) provides a far richer training signal compared to the IU dataset (about 2,000 samples), which explains why MIMIC-trained models consistently outperform IU-trained ones. The VLM-KG (IU) model also improves over LLM-KG (IU) (BLEU-1: 44.76 vs. 25.32), emphasizing the value of multimodal learning, but its lower ROUGE-L score (49.66) indicates that it still struggles with capturing longer contextual dependencies compared to larger datasets.

While Dygiee++ demonstrates strong recall, as reflected in its high ROUGE-L score (56.49), its performance on BLEU metrics is significantly lower than multimodal models. The large difference between BLEU and ROUGE-L scores for Dygiee++ can be attributed to its recall-oriented nature and lack of precision in text generation. ROUGE-L, which measures the longest common subsequence (LCS) between the predicted and reference text, prioritizes recall over precision. This allows Dygiee++ to achieve a high ROUGE-L score (56.49), comparable to LLM-KG (MIMIC) and even surpassing VLM-KG (IU and MIMIC), as it captures key terms from the reference text despite struggling with fluency and structure. In contrast, BLEU relies on exact n-gram matches, meaning that even if the correct entities are present, any variation in phrasing or word ordering significantly lowers the score. This suggests that Dygiee++ struggles to generate outputs that closely match the reference texts in word choice and ordering. Additionally, as a unimodal model relying solely on textual data, Dygiee++ lacks the ability to leverage visual context, which is particularly beneficial in tasks where multimodal reasoning aids in disambiguating complex entity relations.


Figure \ref{fig2} compares the knowledge graphs generated by Dygiee++, VLM-KG (IU), and VLM-KG (MIMIC), demonstrating the impact of multimodal learning on knowledge graph generation. Dygiee++, being a unimodal BERT-based model that processes only the radiology report, produces a dense but noisy graph with redundant and inaccurate relations, as seen in the excessive red-highlighted incorrect entities and relationships. It struggles with entity specificity and often misrepresents key medical concepts, leading to hallucinated relationships as shown in red. In contrast, VLM-KG (IU) significantly reduces errors, with fewer incorrect relations, but still introduces some inaccurate connections (e.g., "acute disease" and "cardiopulmonary"), likely due to limited training data restricting its ability to generalize effectively. VLM-KG (MIMIC), trained on the much larger MIMIC dataset, generates the cleanest and most accurate graph, correctly identifying key relation as shown in black triplets while minimizing incorrect relations. The multimodal nature of the VLM-KG models allows them to better disambiguate clinical terms by incorporating information from both text and radiographic images, leading to more precise and clinically relevant knowledge graphs.

\begin{table}[t]
\caption{Ablation study of the effect of Prefix Tuning. VLM-KG* refers to the setting where the LLM-KG is frozen and only the projector is trained. In VLM-KG both projector and LLM-KG are trained.}
\centering
\begin{tabular}{c|ccccc}
\hline
Model & B1 & B2 & B3 & B4 & RL \\ 
\hline
VLM-KG* & 10.00 & 6.10 & 3.78 & 2.62 & 24.60 \\ 
VLM-KG & \textbf{54.98} & \textbf{49.65} & \textbf{46.12} & \textbf{43.29} & \textbf{54.69} \\ 
\hline
\end{tabular}
\label{tab2}
\end{table}

\begin{table}[t]
\caption{Ablation study of the effect of `clip\_length(k)' and `prefix\_length(n)' in the projector. `clip\_length(k) is the length of the linear layer which reshapes the image feature vector into a sequence format. Prefix is a learned constant added to the sequence. The length of this prefix is given by `prefix\_length(n)'.}
\centering
\begin{tabular}{c|c|ccccc}
\hline
n & k & B1 & B2 & B3 & B4 & RL \\ 
\hline
64  & 64  & \textbf{44.76} & \textbf{38.84} & \textbf{34.84} & \textbf{31.69} & \textbf{49.66} \\ 
128 & 64  & 42.45 & 36.20 & 31.83 & 28.61 & 48.67 \\ 
64  & 128 & 40.24 & 34.35 & 30.52 & 27.67 & 47.19 \\ 
128 & 128 & 38.40 & 32.64 & 28.85 & 26.00 & 46.79 \\ 
\hline
\end{tabular}
\label{tab3}
\end{table}





\begin{table}[t]
\caption{Ablation study of the effect of length of generated tokens.}
\centering
\begin{tabular}{c|ccccc}
\hline
Tokens & B1 & B2 & B3 & B4 & RL \\
\hline
200  & 33.58 & 31.44 & 29.90 & 28.65 & 55.90 \\
256  & 38.66 & 36.10 & 34.21 & 32.65 & \textbf{57.27} \\
300  & \textbf{40.52} & 37.90 & \textbf{35.96} & \textbf{34.35} & 56.52 \\
512  & \textbf{40.52} & \textbf{37.91} & \textbf{35.96} & \textbf{34.35} & \underline{56.53} \\
\hline
\end{tabular}
\label{tab4}
\end{table}

\subsection{Analysis}

We perform ablation studies on our models, bringing more insights into these models. Table \ref{tab2} explores the effect of freezing the LLM-KG in our framework. This is equivalent to prefix tuning. We can see that while prefix tuning may perform well in the general domain, in our case, it performs relatively poor as the task is a specific one, visually instruction tuned LLM-KG performs better in comparison. 

Table \ref{tab3} explores the effect of the prefix length and clip length in the projector of our VLM-KG (IU) model. We see a decrease in the performance of the model when prefix length and clip length are increased. The observed drop in performance with increased prefix length or clip length could be due to information overload or dilution, where the model struggles to focus on key features amidst additional, less relevant data. 

Table \ref{tab4} explores the effect of the length of generated tokens on the quality of generated triplets. Increasing the length of generated tokens increases the BLEU score while the ROUGE-L score increases initially and then drops slightly. We adopted the beam search approach of generation, and we also tested the inference with top-p and top-k sampling, but our model performed better with the beam search strategy.

\subsection{Limitations}

Firstly, our models are limited to the radiology data from MIMIC-CXR and IU-Xray datasets. Secondly, due to data scarcity in radiology the evaluation is limited to small number of radiologist annonated samples. The comparison is limited to single Dygiee++ from \cite{r2} due to unavailability of models from \cite{r2}.

\section{Conclusion}

In this study, we introduced the first multimodal VLM-based framework for knowledge graph generation in radiology, leveraging both textual and visual information to enhance structured knowledge extraction. The integration of multimodal learning enables more accurate triplet generation, reducing hallucinated relations and capturing complex medical concepts more effectively. Our models provide high quality knowledge graph data and can augment many downstream tasks in radiology. We also release the knowledge graph dataset for MIMIC-CXR generated by our multimodal models for advancement of the field of radiology.

%
%
%
%

\end{document}